%% file: main.tex
\documentclass[3p, sort&compress]{elsarticle}

\usepackage[utf8]{inputenc}
\usepackage{markdown}
\usepackage{amssymb}
\usepackage{amsmath}
\usepackage[ruled]{algorithm2e}
\usepackage{multirow}
\usepackage{booktabs}
\usepackage{xcolor}
\usepackage{caption}
\usepackage{subcaption}

\newcommand{\matr}[1]{\mathbf{#1}}

\begin{document}

\input{frontmatter}

\input{sections/introduction}
\input{sections/related_works}
\input{sections/background}
\input{sections/methods}
\input{sections/results}
\input{sections/conclusion}

\clearpage
\bibliographystyle{template/elsarticle-num-names}
\bibliography{main}
\end{document}

%% file: frontmatter.tex
\begin{frontmatter}

\title{Learning-Based Heuristic for Combinatorial Optimization of the Minimum Dominating Set Problem using Graph Convolutional Networks}

\author[1]{Abihith Kothapalli\corref{cor1}}\ead{abi.kothapalli@vanderbilt.edu}
\author[1,2]{Mudassir Shabbir}\ead{mudassir.shabbir@vanderbilt.edu}
\author[1]{Xenofon Koutsoukos}\ead{xenofon.koutsoukos@vanderbilt.edu}

\affiliation[1]{organization={Department of Computer Science, Vanderbilt University}, city={Nashville, TN}, country={USA}}
\affiliation[2]{organization={Department of Computer Science, Information Technology University}, city={Lahore, Punjab}, country={Pakistan}}
\cortext[cor1]{Corresponding author}

\begin{abstract}
\input{sections/abstract}
\end{abstract}


\end{frontmatter}

%% file: sections/abstract.tex
A dominating set of a graph $\mathcal{G=(V, E)}$ is a subset of vertices $S\subseteq\mathcal{V}$ such that every vertex $v\in \mathcal{V} \setminus S$ outside the dominating set is adjacent to a vertex $u\in S$ within the set. The minimum dominating set problem seeks to find a dominating set of minimum cardinality and is a well-established NP-hard combinatorial optimization problem. We propose a novel learning-based heuristic approach to compute solutions for the minimum dominating set problem using graph convolutional networks. We conduct an extensive experimental evaluation of the proposed method on a combination of randomly generated graphs and real-world graph datasets. Our results indicate that the proposed learning-based approach can outperform a classical greedy approximation algorithm. Furthermore, we demonstrate the generalization capability of the graph convolutional network across datasets and its ability to scale to graphs of higher order than those on which it was trained. Finally, we utilize the proposed learning-based heuristic in an iterative greedy algorithm, achieving state-of-the-art performance in the computation of dominating sets.

%% file: sections/introduction.tex
\section{Introduction}

Network-based optimization problems constitute a broad class of problems in the field of combinatorial optimization. These optimization problems offer a means to model highly intricate discrete decision problems across diverse domains where pairwise interactions play a crucial role, such as social network analysis \cite{optim_social_nets}, wireless communications \cite{telecom}, operations research \cite{operations_research}, scheduling \cite{scheduling_optim}, and transportation \cite{transportation_app}. A considerable portion of these problems belongs to the broader class of NP-hard problems, where it is challenging to find exact solutions, as doing so often necessitates a near-complete enumeration of the entire search space. Consequently, computation of exact solutions is practically infeasible, and approximation or heuristic algorithms are generally favored for practical applications. Although these algorithms exhibit significantly faster runtime and possess sub-exponential theoretical complexities, they often yield suboptimal solutions. Therefore, a key area of research revolves around the development of approximation or heuristic algorithms that can provide solutions that are as close to optimal as possible.

The minimum dominating set (MDS) problem is an important network-based optimization problem that involves finding the smallest dominating set of a given graph. A \textit{dominating set} of a graph is a subset of the vertices in the graph such that every vertex is either in the dominating set or adjacent to a vertex in the dominating set. The MDS problem aims to find the dominating set of minimum cardinality. Dominating sets have a wide range of applications in various fields, including social networks \cite{socialnet1, socialnet2, socialnet3}, cybersecurity \cite{cyberapp}, biological networks \cite{bionet}, bioinformatics \cite{bioinformatics}, multi-document summarization \cite{multidoc}, and wireless sensor networks \cite{wirenet} among others. The MDS problem is known to be NP-hard \cite{nphard1, nphard2}. Furthermore, it is also Log-APX-complete, so assuming $\text{P} \neq \text{NP}$, no polynomial-time algorithm can achieve an approximation factor better than $O(\log |\mathcal{V}|)$ for the MDS problem, where $|\mathcal{V}|$ is the number of vertices in the problem instance \cite{apx-hard1, apx-hard2}.

While approximation algorithms can provide theoretical bounds on optimality, these guarantees can be weak or unsatisfactory in general, and these algorithms may have poor empirical performance, if they exist at all \cite{graph-ml, approx}. Alternatively, heuristics lack the theoretical guarantees provided by approximation algorithms but can offer fast algorithms with good empirical performance. However, designing heuristics requires extensive manual trial-and-error and domain expertise \cite{approx, main}. Learning-based approaches have emerged as another viable approach for solving NP-hard problems, leveraging their ability to handle complex problems and learn abstract relationships from large amounts of high-dimensional data. Learning-based approaches can also exhibit faster computation and improved scalability compared to traditional algorithms. Recent works have applied learning-based algorithms to various NP-hard problems, such as maximal independent set, traveling salesman, knapsack, quadratic assignment, minimum vertex cover, and satisfiability \cite{main, tsp, qap, tsp2, graph-ml, sat}. However, these approaches encounter their own challenges, as most problem instances, especially with graph problems, cannot be adequately represented with fixed-length vectors. Additionally, enforcing problem constraints directly on machine learning models can be challenging. There can also be multiple optimal solutions for a given problem instance, requiring an effective learning-based approach to distinguish between distinct nodes in the solution space. Furthermore, NP-hard problems are inherently computationally intractable, and since obtaining labeled training data for these problems necessitates the computation of exact solutions for a series of problem instances, generating a sufficiently large labeled dataset is itself a time-consuming and resource-intensive task.

This work presents a novel graph machine learning framework to compute minimum dominating sets on arbitrary graphs. Given the challenges associated with developing traditional heuristic algorithms, as described earlier, we propose the use of graph convolutional networks (GCNs) to develop a learning-based heuristic. Specifically, our approach employs a GCN to generate a diverse set of likelihood maps over the set of vertices in a given problem instance, and we then treat these probability maps as heuristic functions for use in constructing a dominating set of the graph. We evaluate the empirical performance of the GCN when supplemented with a simple pruning algorithm or implemented in an iterative greedy (IG) algorithm, and compare these results with the state-of-the-art in the computation of dominating sets.

\textbf{Contributions.} The main contributions of this paper can be summarized as follows: 1) We provide a novel dataset of graph instances for the MDS problem with multiple labeled solutions computed per graph instance. 2) We label MDS solutions for graph instances in real-world datasets containing graphs of varied sizes and spanning different settings. 3) We train a GCN model to generate a series of heuristics on input graphs of any arbitrary structure, and we demonstrate that the resulting learning-based heuristics can outperform a classical greedy approximation algorithm. 4) We demonstrate that the GCN model can generalize across datasets and scale to graphs larger than those on which it was trained. 5) We obtain state-of-the-art performance in computation of dominating sets by using the GCN-based heuristics in an IG algorithm. All data and source code required to reproduce our results can be found at \url{https://github.com/abi-kothapalli/MinimumDominatingSets}.

The remainder of this work is structured as follows. Section \ref{sec:related} describes related works in the dominating set literature and graph machine learning. Section \ref{sec:back} then introduces the MDS problem more formally and presents the standard notation and background leveraged throughout the paper, including several key algorithms for the computation of dominating sets. Section \ref{sec:methods} describes our approach to the MDS problem. In Section \ref{sec:results}, we present our empirical results and compare them with the previous state-of-the-art for the MDS problem. Finally, Section \ref{sec:conclusion} concludes this work and summarizes our contributions.

%% file: sections/related_works.tex
\section{Related Works}\label{sec:related}
We briefly review several related works in the literature. Several theoretical works exist that have attempted to bound the size of dominating sets. For a graph $\mathcal{G}$ with $n$ vertices, let the minimum and maximum degree of any vertex in $\mathcal{G}$ be $\delta$ and $\Delta$, respectively, and let $d$ be the diameter of the graph (that is, the maximum number of edges on the shortest path between any two vertices in $\mathcal{G}$). We denote by $\gamma(\mathcal{G})$ the domination number of $\mathcal{G}$, which is simply the size of the smallest dominating set of $\mathcal{G}$. It has been shown that $\gamma(\mathcal{G})$ satisfies both the bounds $\frac{n}{\Delta + 1} \leq \gamma(\mathcal{G}) \leq \frac{n}{2}$ and $\frac{d+1}{3} \leq \gamma(\mathcal{G}) \leq n - \Delta$ \cite{theory1,theory2}. Moreover, \cite{prob} show that if $\delta > 1$, then $\gamma(\mathcal{G}) \leq n\frac{1+\ln(\delta + 1)}{\delta + 1}$. For further discussion on the tightest known bounds on $\gamma(\mathcal{G})$ for various values of $\delta$, we direct the reader to \cite{bounds}. While such bounds provide useful theoretical results, it remains intractable to find optimal solutions to the MDS problem on general graphs.

Exact algorithms for the MDS problem have also been studied extensively. To the best of our knowledge, the current best exact algorithm for the MDS problem is presented in \cite{exact}. They employ a branch and reduce based algorithm to compute exact solutions to the MDS problem. Using a measure and conquer approach, the authors determine the runtime complexity of their algorithm to be $O(1.4969^n)$ while only requiring polynomial space. Faster algorithms for specific subclasses of graphs have also been developed. In \cite{subgraphs}, the authors discuss exact algorithms for chordal graphs, circle graphs, and dense graphs, which provide improvements in runtime compared to the $O(1.4969^n)$ complexity required for general graphs. However, these algorithms are still exponential in complexity. Meanwhile, linear time algorithms for series-parallel graphs, $k$-degenerated graphs, and trees are presented in \cite{series}, \cite{degen}, and \cite{tree}, respectively.

When we are not restricted to specific subclasses of graphs, however, the computation of minimum dominating sets remains intractable. As a result, there is significant interest in heuristic and approximation algorithms for the MDS problem. The most well-known approximation algorithm for the MDS problem uses a greedy heuristic that iteratively adds the vertex with the greatest number of non-dominated neighbors to a set until that set forms a valid dominating set of the graph. In \cite{greedy}, it is shown that the size of the set returned by this algorithm is upper-bounded by $n+1-\sqrt{2m+1}$, where $n:=|\mathcal{V}|$ and $m:=|\mathcal{E}|$ represent the number of vertices and edges, respectively, in the problem instance. This algorithm also achieves an $O(\log |\Delta|)$ approximation factor, and \cite{approx-factor} demonstrate that, in fact, the logarithm approximation factor is the best one can do, assuming $\text{P} \neq \text{NP}$. Variants of this greedy heuristic and their empirical performances are discussed in \cite{heuristics}.

Conversely, several heuristic algorithms exist in the literature which do not offer the same theoretical guarantees as approximation algorithms but demonstrate strong empirical performance. We will briefly discuss the state-of-the-art algorithms for the computation of dominating sets. The most recent is an iterative greedy (IG) algorithm, proposed in \cite{iterative}, which constructs an initial dominating set and iteratively destructs and reconstructs portions of the solution to improve the solution size. The randomized local search (RLS) algorithm, presented in \cite{rls}, builds solutions from different permutations of the vertices in a problem instance using a greedy approach and incorporates a so-called jump operator to enhance the solutions. Finally, \cite{aco} presents an ant colony optimization (ACO) algorithm enhanced with local search. This method generates populations of solutions randomly, which then evolve probabilistically while using local search to prune out redundant vertices. Variants of the ACO algorithm are presented in \cite{rls} and \cite{aco-v2}. In \cite{iterative}, the empirical performance of all these methods is compared, and it is shown that the IG algorithm outperforms the others. Therefore, we primarily benchmark the performance of our proposed algorithms against this IG algorithm. Further details on the IG algorithm are provided in Section \ref{sec:ig}.

Finally, we discuss related advances in graph machine learning that we leverage in this work. Specifically, we use the GCN architecture, a type of graph neural network, originally introduced in \cite{kipf2016semi}. In our approach, we draw inspiration from \cite{main}, which demonstrates the application of the GCN architecture to generate solutions for combinatorial optimization problems on graphs. In their setup, the GCN architecture is trained to learn a diverse set of probability maps over problem instances, which are then used in a tree search to obtain solutions to various combinatorial optimization problems, namely the satisfiability, maximal independent set, minimum vertex cover, and maximal clique problems. In our work, we posit that the probability maps learned by the GCN can directly serve as a diverse set of learning-based heuristic functions for the combinatorial optimization problem at hand. For further discussion on the use of graph machine learning for combinatorial optimization, we direct the reader to \cite{ijcai2021p595}.

%% file: sections/background.tex
\section{Background}\label{sec:back}
Let $\mathcal{G}=(\mathcal{V, E})$ be a simple graph where $\mathcal{V}$ represents the set of vertices in $\mathcal{G}$ and $\mathcal{E}$ represents the set of edges. For a given vertex $v \in \mathcal{V}$, we define the open neighborhood of $v$, denoted as $N(v)$, as the set $\{u \in \mathcal{V} : (u, v) \in \mathcal{E}\}$. Similarly, the closed neighborhood of $v$, denoted as $N[v]$, is defined as $N(v) \cup \{v\}$. We can extend these definitions to sets of vertices $S\subseteq \mathcal{V}$ such that $N(S) := \bigcup_{v\in S} N(v)$ is the open neighborhood of $S$, and $N[S] := \bigcup_{v\in S} N[v] = N(S) \cup S$ is the closed neighborhood of $S$.

A set $S\subset \mathcal{V}$ is considered a \textit{dominating set} of $\mathcal{G}$ if and only if the closed neighborhood of $S$ spans the vertex set of $\mathcal{G}$, i.e., $N[S] = \mathcal{V}$. Equivalently, $S$ is a dominating set of $G$ if for every $v\in \mathcal{V}$, $v\in S$ or $\exists \; u \in S$ such that $(u, v)\in\mathcal{E}$. That is, every vertex in $\mathcal{V}$ is either in $S$ or adjacent to a vertex in $S$. In the \textit{minimum dominating set (MDS)} problem, we seek a dominating set, $S^*$, of minimum cardinality, i.e., $|S^*|\leq |S|$, for all valid dominating sets $S$ of $\mathcal{G}$. It is important to note that for certain graphs, $S^*$ may not be unique, as there may exist multiple solutions to the MDS problem, each with the same cardinality. The cardinality of the minimum dominating set is referred to as the domination number of $\mathcal{G}$, denoted as $\gamma(\mathcal{G})$.

\subsection{Integer Linear Programming Formulation}\label{sec:ilp}

One effective and practical approach for challenging combinatorial optimization problems is through an integer linear programming (ILP) formulation. Although the equivalent ILP problem remains NP-hard, there exist highly optimized standard linear programming solvers that can efficiently solve small to moderate-sized instances of these problems. The MDS problem can also be reduced to an ILP problem, and we use this formulation in Section \ref{sec:data} to compute exact solutions to the MDS problem. 

The ILP formulation for the MDS problem is as follows. For a graph $\mathcal{G = (V, E)}$, where $\mathcal{V} = \{v_i\}_{i=1}^n$ is the set of $n$ vertices in the graph, we define a binary variable $x_i\in \{0, 1\}$ for each vertex $v_i$. This binary variable indicates whether the corresponding vertex is included in the MDS solution. Specifically, $x_i$ is set to $1$ if $v_i$ is selected to be included in the MDS, and $x_i = 0$ otherwise. We then define the objective function of the MDS problem as:
\begin{equation}\label{eq:ilp}
    \min \quad \sum_{i=1}^n x_i.
\end{equation}
To ensure that a solution is a valid dominating set, it must satisfy the following constraints:
\begin{equation}\label{eq:constraint1}
    \sum_{\substack{j\\v_j\in N[v_i]}} x_j \geq 1 \qquad \forall i = 1, \dots, n.
\end{equation}
Alternatively, if $\matr{A} = \begin{bmatrix}a_{ij}\end{bmatrix}\in \{0, 1\}^{n\times n}$ is the symmetric adjacency matrix for $\mathcal{G}$ such that $a_{ij} = 1$ if $(v_i, v_j) \in \mathcal{E}$ and $a_{ij} = 0$ otherwise, we can represent Equation \ref{eq:constraint1} equivalently as:
\begin{equation}\label{eq:constraint2}
    x_i + \sum_{j=1}^n a_{ij} x_j \geq 1 \qquad \forall i =1, \dots n.
\end{equation}
These constraints ensure that the selected vertices indeed form a valid dominating set for $\mathcal{G}$ by enforcing that for every vertex, either the vertex itself is selected or at least one of its adjacent vertices is selected. The objective function in Equation (\ref{eq:ilp}) minimizes the number of selected vertices, ensuring the optimality of the resulting solution. It is worth noting that in the resulting solution, $\gamma(\mathcal{G}) = \sum_{i=1}^n x_i$, which is exactly the objective function being optimized.

\subsection{Heuristic Approaches}\label{sec:heuristic}

The exact solutions to the minimum dominating set problem, including the integer programming formulation mentioned above, have exponential time complexity and do not scale well. Therefore, various heuristic approaches to find efficient solutions that may not be optimal have been developed. We outline the general structure of a greedy heuristic approach for the MDS problem in Algorithm \ref{alg:heuristic}. In this algorithm, $h: \mathcal{V} \to \mathbb{R}$ defines a real-valued heuristic function, and vertices that maximize $h(\cdot)$ are greedily selected until a valid dominating set is formed.

\begin{algorithm}
\caption{A greedy heuristic algorithm to find a dominating set for $\mathcal{G}$}\label{alg:heuristic}
\DontPrintSemicolon
\KwData{$\mathcal{G} = (\mathcal{V, E})$}
\KwResult{$S \subseteq \mathcal{V}$, that dominates $\mathcal{G}$}
\Begin{
    $S\gets \{\}$\;
    
    \While{$S$ does not dominate $\mathcal{G}$}{
        $S'\gets\left\{v : h(v) = \max\limits_{u\in\mathcal{V}\setminus S}\{h(u)\}\right\}$\;
        $S \gets S \cup S'$\;
    }
    \KwRet{$S$}
}
\end{algorithm}

We use two different traditional heuristics as baselines for performance comparisons. The first heuristic corresponds to a well-known greedy approximation algorithm for the MDS problem. This heuristic counts the number of \textit{non-dominated} neighbors of a given vertex, i.e., the subset of neighbors not dominated by the current dominating set under construction, denoted as $S$. The greedy heuristic function, which we denote as $h_{g}(v)$, is defined as follows:
\begin{equation}\label{eq:greedy}
    h_{g}(v) = \left|N[v] \setminus N[S] \right|.
\end{equation}
When this heuristic is used, we prioritize adding vertices to the solution that will dominate the greatest number of previously non-dominated vertices in the graph. This heuristic is an analog of a greedy heuristic algorithm originally introduced in \cite{greedy_set_cover} for the set cover problem, but it has since been adapted to a variety of related hard combinatorial optimization problems, including the vertex cover problem and the MDS problem itself \cite{greedy_set_cover_2, greedy}. As mentioned previously, the use of this particular heuristic is equivalent to a classical greedy approximation algorithm for the MDS problem, and it thus provides certain theoretical guarantees on its performance \cite{greedy}.

The second heuristic we use is a random heuristic, which assigns a random value to each vertex in the graph:
\begin{equation}\label{eq:rand}
    h_{r}(v) \sim \mathcal{U}(0,1).
\end{equation}
As a result, when $h_r$ is used in Algorithm \ref{alg:heuristic}, vertices will be added to the solution set $S$ in a random order until the set $S$ dominates $\mathcal{G}$. This approach represents a naive strategy for constructing a dominating set, as it does not consider any information from the topology of the input graph. We include this heuristic primarily for illustrative purposes, as it serves as a baseline for comparison with other, more informed heuristics.

For any optimal solution to an instance of the MDS problem, there must exist a valid optimal priority ordering of the vertices in the input graph that leads to the construction of the optimal solution, using a selection procedure similar to Algorithm \ref{alg:heuristic}. However, there is no tractable algorithm that can determine this ordering a priori, and as such, heuristic functions simply seek an ordering that minimizes the size of the resulting dominating set. Given the recent success of data-driven methodologies, it is natural to explore a machine learning-based approach that, given a set of MDS problem instances and their corresponding solutions, learns an optimal heuristic function. Nevertheless, such an approach does face its own challenges, and implementing off-the-shelf machine learning models may fail to yield satisfactory results. Firstly, an MDS problem instance takes the form of a graph, which is an irregular and permutation-invariant data structure. Such a data structure is not readily compatible with most machine learning models that expect input in the form of vectors in fixed-dimensional Euclidean space. Additionally, an MDS problem instance may not admit a unique solution; in fact, many MDS problem instances have an exponential number of optimal solutions, which can make a model prone to daunting error rates. In the following section, we attempt to address these challenges and train a GCN model to generate learning-based heuristic functions that we can use in the above-described algorithm. The GCN model is trained to compute probability maps over the vertex set $\mathcal{V}$ of the input graph, predicting the probability of each vertex being part of an optimal solution. We can then directly employ these probabilities as a heuristic function.

\begin{figure}[tb]
     \centering
     \begin{subfigure}[b]{0.3\columnwidth}
         \centering
         \includegraphics[width=\columnwidth]{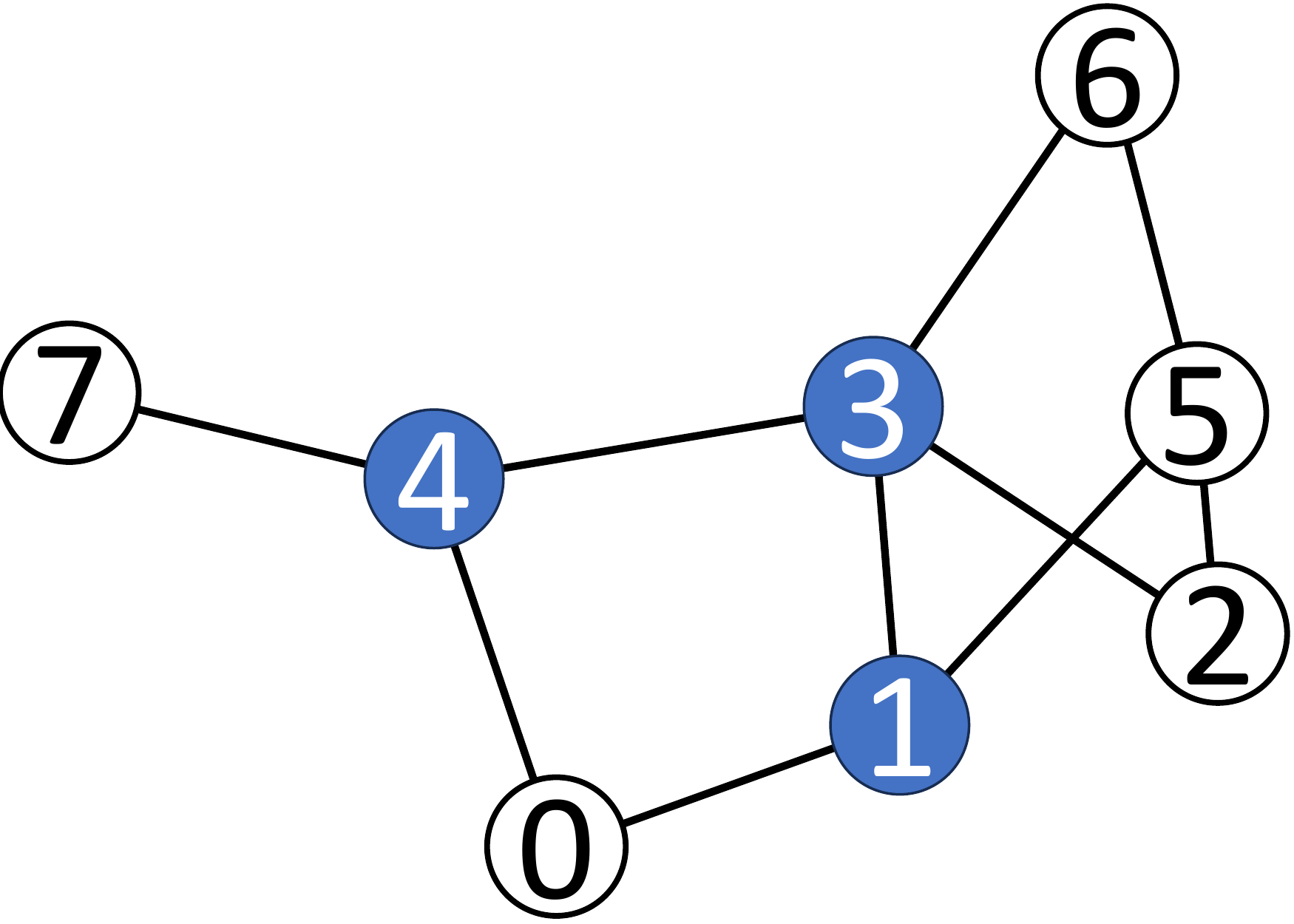}
         \caption{Dominating set computed with classical greedy heuristic (Equation \ref{eq:greedy}). This heuristic selects the vertices in the following order: $3\to 1 \to 4$.}
         \label{fig:heuristic-a}
     \end{subfigure}
     \hfill
     \begin{subfigure}[b]{0.3\columnwidth}
         \centering
         \includegraphics[width=\columnwidth]{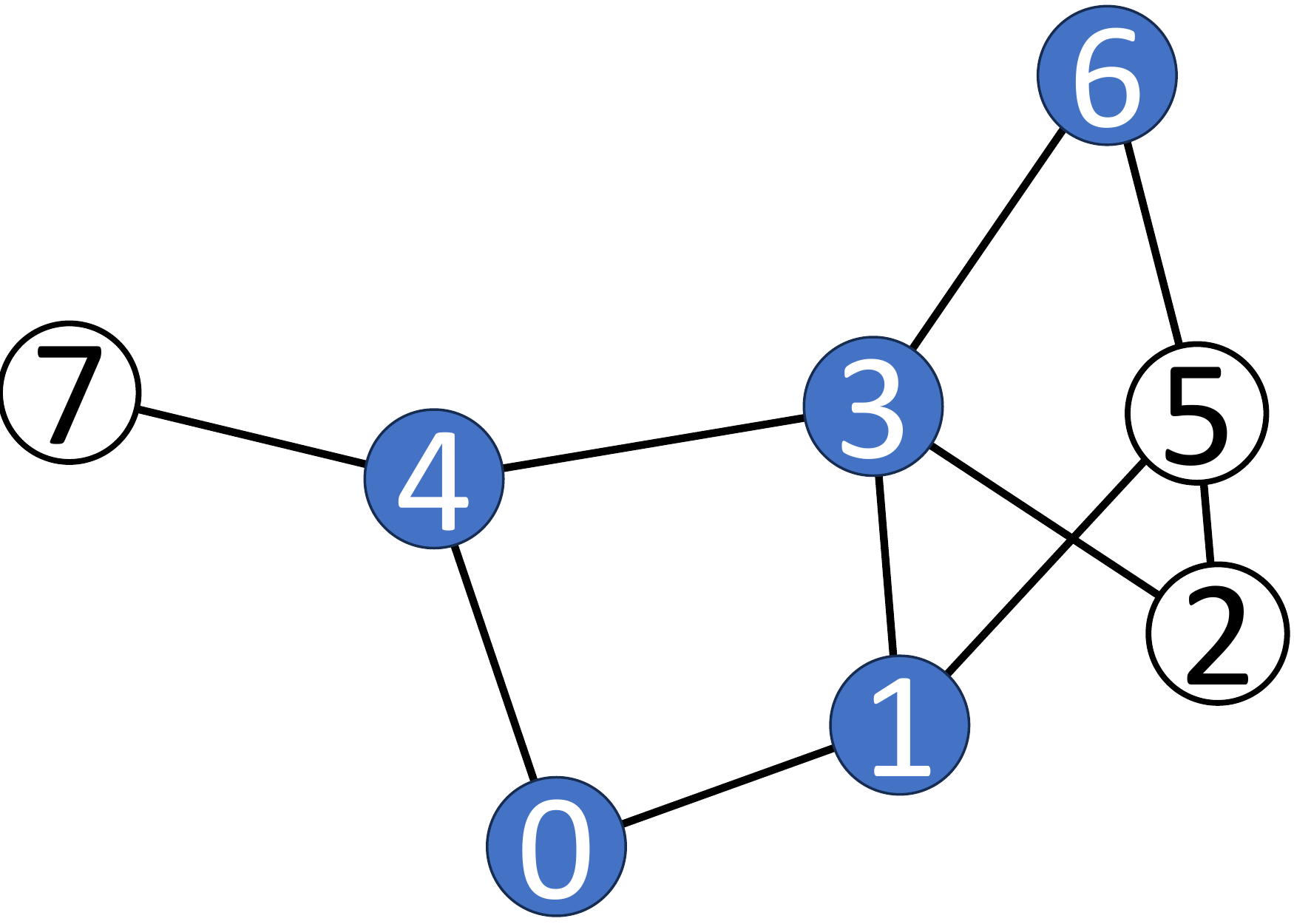}
         \caption{Dominating set computed with random heuristic (Equation \ref{eq:rand}). In this example, the vertices were selected in the following order: $3\to 6\to 1\to 0\to 4$.}
         \label{fig:heuristic-b}
     \end{subfigure}
     \hfill
     \begin{subfigure}[b]{0.3\columnwidth}
         \centering
         \includegraphics[width=\columnwidth]{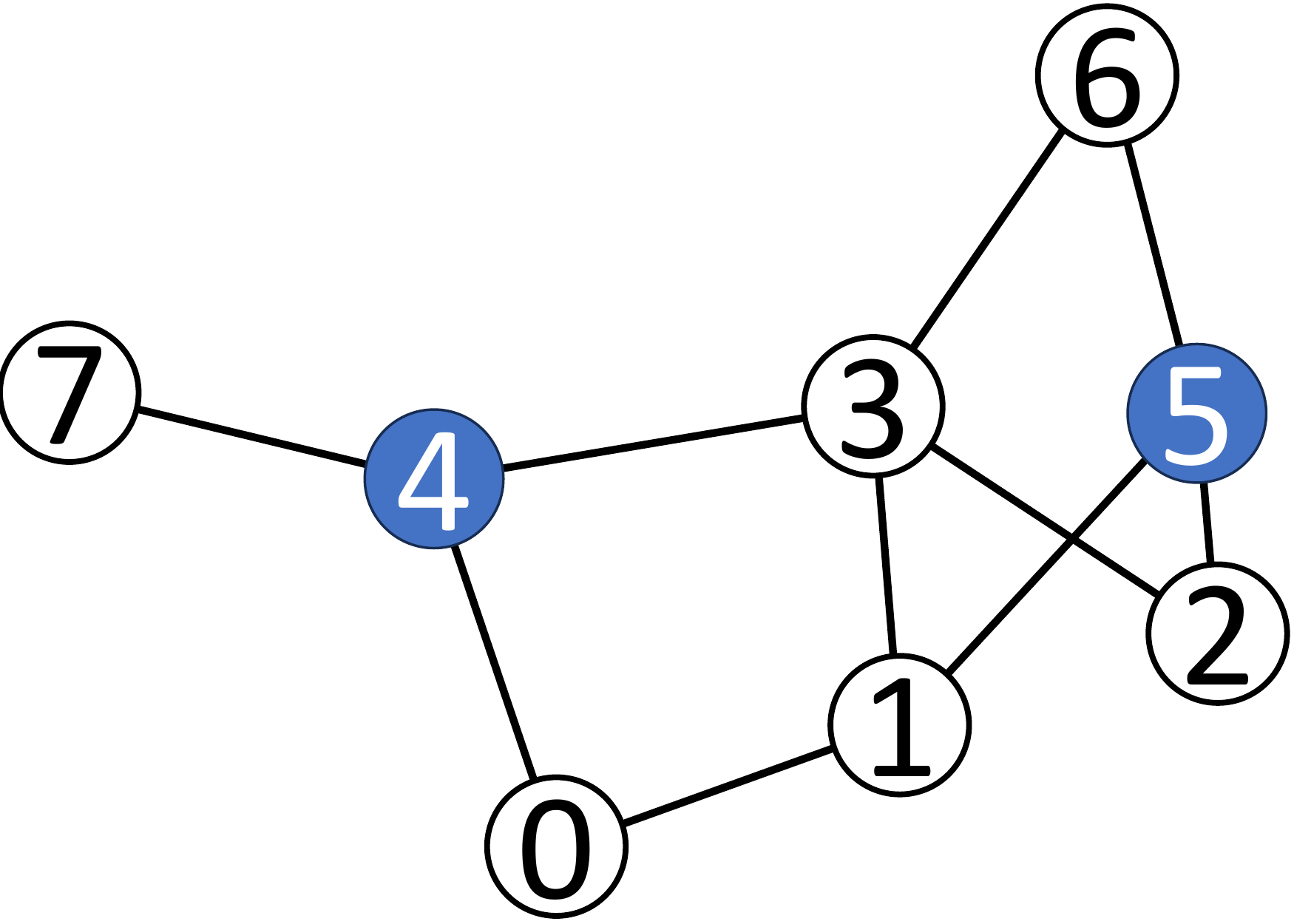}
         \caption{Optimal MDS.}
         \label{fig:heuristic-c}
     \end{subfigure}
     
    \caption{Examples of dominating sets computed by heuristic approaches. Note that Figure \ref{fig:heuristic-b} is just one possible dominating set computed by the random heuristic, as the heuristic is not deterministic.}
    \label{fig:heuristic}
\end{figure}


\subsection{Iterative Greedy Algorithm}\label{sec:ig}

Finally, we provide a brief overview of the IG algorithm as presented in \cite{iterative}. IG is a versatile, hybrid metaheuristic framework that can be applied toward a variety of problem domains and shares similarities with other popular metaheuristic methods such as simulated annealing and tabu search. One of its first applications was in solving a type of scheduling problem, as outlined in \cite{ig}. It has since been successfully utilized for other NP-hard problems, including the binary quadratic programming problem \cite{ig_quad} and the traveling salesman problem \cite{ig_tsp}. For a more comprehensive understanding of the IG framework, we refer readers to \cite{ig_rev}. To our knowledge, \cite{iterative} were the first to adapt this framework to the MDS problem and demonstrate that their IG algorithm could achieve state-of-the-art performance in computation of dominating sets.

The pseudocode for the IG algorithm to compute dominating sets is given in Algorithm \ref{alg:ig}. The algorithm begins by generating an initial valid dominating set using Algorithm \ref{alg:heuristic} with the greedy heuristic function $h_g(v)$ from Equation \ref{eq:greedy}. It then applies a local search procedure to further refine the solution. Then, the algorithm iteratively destructs and reconstructs the dominating set, incorporating the local search procedure after each reconstruction. The destruction phase takes an input parameter $\beta$, which specifies the proportion of the dominating set to randomly be destroyed. The reconstruction phase then greedily adds back vertices to the partially destructed set until it is once again a valid dominating set, employing the same greedy heuristic $h_g(v)$. For a more detailed explanation of each stage, see \cite{iterative}. We also note that the algorithm also takes an input parameter $\Delta$ that limits the number of iterations without improvement, but in practice, we also impose a time limit on the algorithm, similarly to \cite{iterative}.

\begin{algorithm}
\caption{Iterative greedy (IG) algorithm to compute dominating set}\label{alg:ig}
\DontPrintSemicolon
\KwData{$\mathcal{G} = (\mathcal{V, E}), \beta, \Delta$}
\KwResult{$S \subseteq \mathcal{V}$, that dominates $\mathcal{G}$}
\Begin{
    $S_0 \gets InitialSolution(\mathcal{G})$\;
    $S^* \gets LocalImprovement(S_0)$\;
    $\delta \gets 0$\;
    \While{$\delta < \Delta$}{
        $S'_d \gets RandomDestruction(S^*, \beta)$\;
        $S'_r \gets Reconstruction(S'_d)$\;
        $S' \gets LocalImprovement(S'_r)$\;

        \If{$|S'| < |S^*|$}{
            $S^* \gets S'$\;
            $\delta \gets 0$\;
        }
        \Else{
            $\delta \gets \delta + 1$\;
        }
    }
    \KwRet{$S^*$}
}
\end{algorithm}

Since this approach uses the classical greedy heuristic given by $h_g(v)$ in its $InitialSolution$ and \linebreak $Reconstruction$ procedures, we hypothesize that we could instead use the heuristic learned by the GCN in these procedures. As we later discuss in further detail, the GCN is trained to learn multiple diverse probability maps, and since these probability maps can be treated as heuristic functions, we can iterate through these different probability maps during the IG procedure, enabling the algorithm to leverage the diversity of the heuristic functions learned by the GCN. This has the potential to yield higher-quality solutions compared to those obtained using the classical greedy heuristic alone.

%% file: sections/methods.tex
\section{Methodology}\label{sec:methods}

The task of devising an effective heuristic that can accurately approximate optimal solutions for the MDS problem is an intricate endeavor, demanding substantial iterative refinement and domain expertise. In this section, we present a novel methodology striving to acquire MDS-specific insights from an extensive collection of problem instances and their corresponding solutions. By adopting a data-driven, learning-based approach, we aim to surpass the performance of existing heuristics, offering a promising avenue for addressing the challenges posed by the MDS problem. Formally, we train a specialized neural network, denoted as $f(\cdot)$, which takes a graph $\mathcal{G = (V, E)}$ as input and produces a set of probability maps over $\mathcal{V}$. Each probability map $\hat{y} \in [0,1]^{n}$ indicates the likelihood of each individual vertex belonging to an optimal MDS solution. Such a probability map can then be used to define a heuristic $h(v_i) = \hat{y}_i$, where $\hat{y}_i$ is the output probability corresponding to vertex $v_i$, enabling us to construct dominating sets as described in Algorithm \ref{alg:heuristic}. In the following, we discuss the design and development of this neural network, starting with the essential task of dataset generation.

\subsection{Dataset Generation}\label{sec:data}
The generation of high-quality datasets plays a critical role in solving combinatorial optimization problems using supervised learning techniques. These problems often involve discrete decision-making processes, and the quality of the data directly impacts the ability of the learning algorithm to capture underlying patterns and relationships effectively. Therefore, carefully crafting and selecting datasets that accurately represent the problem space and capture the relevant information can lead to improved model performance and better optimization results. Additionally, the dataset should encompass a diverse range of problem instances to ensure the model generalizes well to unseen data. In the context of the MDS problem, an additional challenge arises from the existence of multiple optimal solutions for a single instance, and training a machine learning model on only one of those solutions would be insufficient to achieve acceptable accuracy. 

While existing literature provides instances of graphs for the MDS problem (e.g., see \cite{iterative, rls}), they lack the domination number and solutions for the corresponding instances, let alone multiple labeled sets of vertices that comprise different optimal solutions for each input graph. In the following, we propose to generate our own dataset of instances for the MDS problem and compute optimal dominating sets for each instance to serve as labeled training data. Our dataset comprises a total of 1349 random binomial graphs generated using the Erd\H{o}s–R\'{e}nyi model with varying orders and edge densities. The size of the graphs ranges from 150 to 255 vertices, with an average size of 192 vertices. The average dominination number of the graphs in this dataset is 25 vertices. It is worth noting that we choose to generate relatively sparser graphs, as the MDS for dense graphs tends to consist of fewer vertices and is generally easier to compute. The optimal solutions for each instance are computed using the ILP approach discussed below.

To compute the optimal dominating sets for each instance in the dataset, we reduce the MDS problem on each graph to an equivalent ILP problem, as described in Section \ref{sec:ilp}. The ILP problem can then be solved using a linear programming optimizer, such as those provided by \cite{pulp, coin}. Once the optimization problem is solved, we define the solution set for the MDS problem instance (using the notation from Section \ref{sec:ilp}) as $S = \{v_{k_1}, v_{k_2}, \dots, v_{k_\gamma}\}$ where $\gamma(\mathcal{G}) = \sum_{i=1}^n x_i$ and $x_{k_i} = 1$ for $i \in \{1, \dots, \gamma\}$. This procedure yields a single optimal dominating set per instance in the dataset. However, as noted previously, the MDS for a given instance is not always unique. Therefore, to obtain multiple diverse optimal solutions for each instance, we introduce an additional constraint to the optimization problem and solve the modified problem. The additional constraint is defined as follows:
\begin{equation}\label{eq:extra}
    \sum_{i=1}^\gamma x_{k_i} \leq \gamma - 1.
\end{equation}
By adding this additional constraint, we force the ILP solver to generate another MDS for the input graph that is distinct from the previous solution. We can then repeat this process by adding another contraint equation given by the newly obtained solution, yielding a third MDS solution, and so forth. It is important to note that the size of the solution set is checked after each iteration to ensure that the additional constraint defined by Equation \ref{eq:extra} does not change the size of the solution set. That is, for the updated set of binary labels $\{x_i\}_{i=1}^n$, we check that $\sum_{i=1}^n x_i = \gamma(\mathcal{G})$. This procedure ultimately allows us to generate multiple optimal solutions for each instance, capturing the diverse nature of optimal dominating sets for the MDS problem.

\subsection{Training GCN Model}\label{sec:gcn}
The input for the MDS problem is a graph, and therefore, traditional machine learning models that take regular vectorized input cannot be used. Hence, we use a specialized graph neural network architecture for this purpose. We train our network $f(\cdot)$, adapting the GCN architecture presented in \cite{main}. To train the model, we use a subset of graphs from the synthetically-generated dataset described in Section \ref{sec:data}. We outline the details of the architecture below. 

Let $\mathcal{D}=\{(\mathcal{G}_i, \mathcal{I}_i)\}$ be the training set, where $\mathcal{I}_i \in \{0, 1\}^{n_i}$ is a binary representation of one of the optimal MDS solutions generated for the graph instance $\mathcal{G}_i$ with $n_i$ vertices. That is, for each vertex $v_j$, a one in the solution vector $\mathcal{I}_i$ reflects the fact that $v_j$ is part of the MDS, and a zero reflects otherwise. The network $f(\mathcal{G}_i; \boldsymbol{\theta})$ is parameterized by $\boldsymbol{\theta}$ and is trained to produce $m$ probability maps:
\begin{equation*}
    \left\langle f^1(\mathcal{G}_i; \boldsymbol{\theta}), f^2(\mathcal{G}_i; \boldsymbol{\theta}), \dots, f^m(\mathcal{G}_i; \boldsymbol{\theta})\right\rangle.
\end{equation*}
Each probability map $f^k(\mathcal{G}_i; \boldsymbol{\theta}) \in [0, 1]^{n_i}$ encodes the likelihood of each vertex in $\mathcal{G}_i$ belonging to an optimal MDS solution. 

\begin{figure}[tb]
    \centering
    \includegraphics[width=0.8\columnwidth]{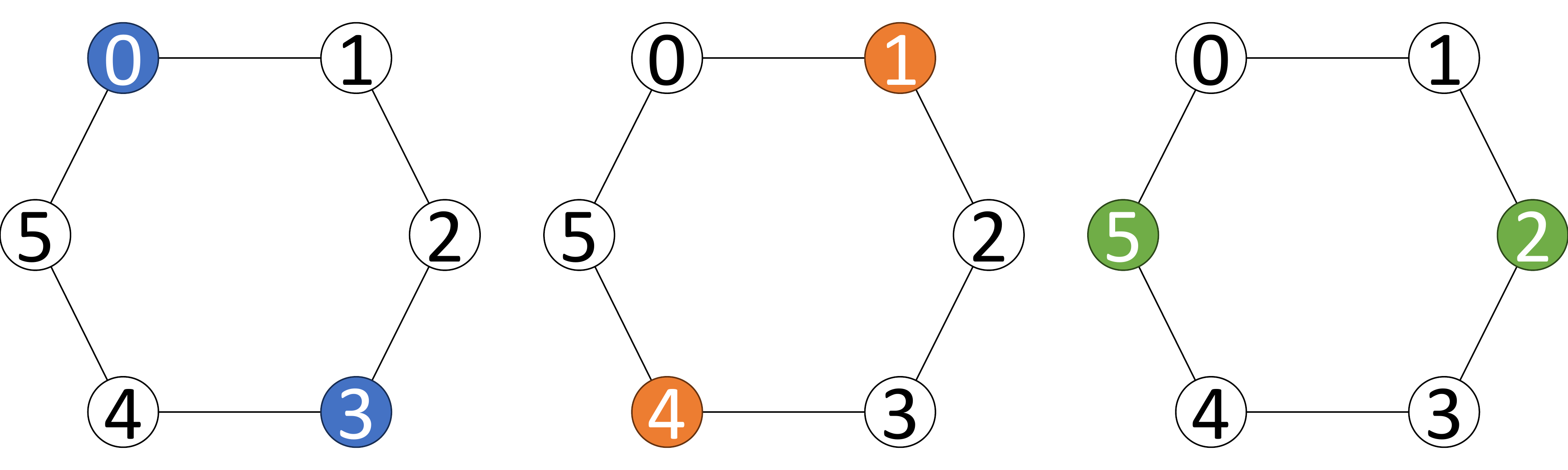}
    \caption{Example of non-uniqueness of MDS solutions. This figure illustrates the following three unique MDS solutions for the same graph: $\{0, 3\}, \{1, 4\}, \{2, 5\}.$}
    \label{fig:mds}
\end{figure}

The rationale behind generating multiple probability maps is to capture the diversity of solutions that can exist for a given instance of the MDS problem. Since there can often exist several different and non-overlapping solutions to a given problem instance, a network that outputs only a single probability map may not capture the full range of possible solutions. Consider for example Figure \ref{fig:mds}, which illustrates a graph with three unique MDS solutions that are entirely non-overlapping. A naively designed network architecture might produce a probability map that assigns equal likelihood to each vertex. This would not provide a useful heuristic, as it is effectively the same as the random heuristic. To overcome this limitation, our goal is to generate multiple high-quality probability maps that can be used to generate diverse solutions for a single input graph. By training the network on a diverse dataset of instances with multiple labeled solutions, we aim to enable the network to learn and generate a range of probability maps that capture the various valid dominating sets for different graph instances.

These probability maps are generated via an adaptation of the GCN architecture originally proposed in \cite{kipf2016semi}. The GCN consists of $L+1$ layers $\{\mathbf{H}^l\}_{l=0}^L$, where $\mathbf{H}^l\in\mathbb{R}^{n\times C^l}$ is the $l^{th}$ feature layer, $C^l$ is the number of feature channels in the $l^{th}$ layer, and $n$ is the number of vertices in the input graph. Since the network receives a graph $\mathcal{G}$ without any vertex-specific feature vectors as input, we let $\mathbf{H}^0 = \mathbf{1}_{n, C^0}$, meaning that $\mathbf{H}^0$ contains rows of all-one vectors of size $C^0$. This ensures that the network treats all vertices equivalently, and predictions are made solely based on the structure of the graph. Each subsequent layer $\mathbf{H}^{l+1}$ is then computed from the previous layer as follows:
\begin{equation}
    \mathbf{H}^{l+1} = \sigma(\mathbf{H}^l\boldsymbol{\theta}_0^l + \mathbf{\Gamma}^{-\frac{1}{2}}\mathbf{A}\mathbf{\Gamma}^{-\frac{1}{2}}\mathbf{H}^l\boldsymbol{\theta}_1^l)
\end{equation}
where $\matr{A}\in\{0, 1\}^{n\times n}$ is the symmetric adjacency matrix of the graph; $\boldsymbol{\theta}_0^l, \boldsymbol{\theta}_1^l \in \mathbb{R}^{C^l\times C^{l+1}}$ are the layer-specific trainable weight matrices; $\mathbf{\Gamma}$ is the diagonal vertex degree matrix of $\mathbf{A}$ with diagonal entries $\mathbf{\Gamma}_{i, i}=\deg(v_i)$, and $\mathbf{\Gamma}^{-\frac{1}{2}}\mathbf{A}\mathbf{\Gamma}^{-\frac{1}{2}}$ is the symmetric normalization of $\mathbf{A}$; and $\sigma(\cdot)$ is a nonlinear activation function. For the final output layer $\mathbf{H}^L$ we use the sigmoid activation function, and for all other layers we use ReLU. In the output layer, we set $C^L=m$, and we treat each row in the output layer as a probability map: $\mathbf{H}^L_k = f^k(\mathcal{G}; \boldsymbol{\theta})$. Here, we use $m$ to refer to the total number of output probability maps and $k$ as an index to an arbitrary output probability map.

We train the network to optimize the hindsight loss on the training set, defined as:
\begin{equation}\label{eq:loss}
    \mathcal{L}(\mathcal{D}, \boldsymbol{\theta}) = \sum_i \min_k \ell(\mathcal{I}_i, f^k(\mathcal{G}_i; \boldsymbol{\theta}))
\end{equation}
where
\begin{equation}
    \ell(\mathcal{I}_i, f^k(\mathcal{G}_i; \boldsymbol{\theta})) = \sum_{j=1}^n \Big[ \mathcal{I}_{ij}\log f_j^k(\mathcal{G}_i; \boldsymbol{\theta}) + (1-\mathcal{I}_{ij})\log(1-f_j^k(\mathcal{G}_i; \boldsymbol{\theta})) \Big]
\end{equation}
is the binary cross-entropy loss for a single given probability map. Here, $\mathcal{I}_{ij}$ denotes the $j^{th}$ element of $\mathcal{I}_i$ and similarly, $f_j^k(\mathcal{G}_i; \boldsymbol{\theta})$ is the $j^{th}$ element of $f^k(\mathcal{G}_i; \boldsymbol{\theta})$. The hindsight loss used here ensures that the loss for each training sample is determined only by the best of the $m$ different probability maps produced by the network; this encourages the network to develop highly diverse probability maps that can ultimately allow us to construct a variety of unique solutions.

\subsection{Using GCN Model to Construct Dominating Sets}\label{sec:output}
\begin{figure}[h]
    \centering
    \includegraphics[width=\columnwidth]{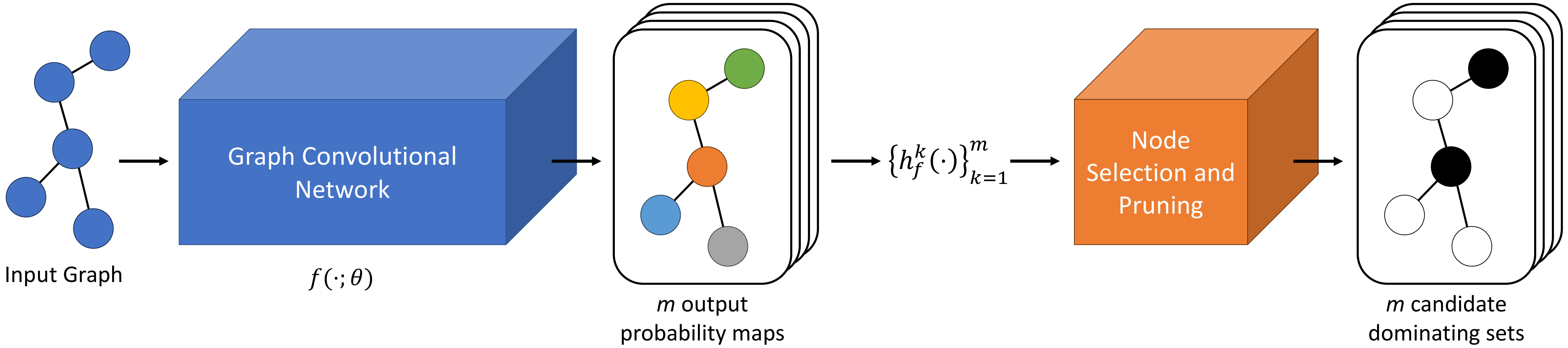}
    \caption{Illustration of pipeline to construct dominating sets using GCN. For a given input graph, the GCN $f(\cdot; \theta)$ outputs a set of $m$ probability maps, which we treat as $m$ different heuristic functions $\{h_f^k(\cdot)\}^m_{k=1}$. We use each of these heuristic functions in the selection procedure given by Algorithm \ref{alg:heuristic} to construct a dominating set, and then prune each of these dominating sets with Algorithm \ref{alg:prune}. Hence, $m$ different candidate dominating sets are constructed, from which we can choose the solution of minimum cardinality.}
    \label{fig:methods}
\end{figure}

To construct a collection of $m$ dominating sets for a test graph instance $\mathcal{G}$ using the trained network $f(\cdot; \boldsymbol{\theta})$, we generate a set of $m$ heuristic functions $\{h_f^k(\cdot)\}^m_{k=1}$ based on the probability maps output by the network. Specifically, for the $k^{th}$ probability map $\hat{y} = f^k(\mathcal{G}_i; \boldsymbol{\theta})$, we define a heuristic $h^k_f(v_i) = \hat{y}_i$, as described previously. We can then directly employ each of these heuristics to generate a dominating set using Algorithm \ref{alg:heuristic}. Since we have $m$ probability maps for each input graph $\mathcal{G}$, we can define $m$ different heuristic functions $\{h_f^k(\cdot)\}^m_{k=1}$, which ultimately result in $m$ different candidate dominating sets. We illustrate this process in Figure \ref{fig:methods}.

We also apply a pruning algorithm to refine each of the constructed dominating sets. This pruning algorithm is given in Algorithm \ref{alg:prune}. We find that since Algorithm \ref{alg:heuristic} greedily chooses vertices using a given heuristic, the resulting dominating sets often end up containing redundant vertices, regardless of the exact heuristic being employed. However, we can easily identify vertices that can be safely removed from the dominating set using a greedy approach, thus minimizing the solution size. Once each of the $m$ candidate dominating sets are pruned, we choose the set with minimum cardinality as our solution. We will hereafter refer to this setup simply as the ``GCN'' algorithm and the classical greedy and random heuristic approaches given in Section \ref{sec:heuristic} as the ``Greedy'' and ``Random'' algorithms, respectively. Note that we will also apply this pruning algorithm in the Greedy and Random algorithms, in order to accurately compare performance.

\begin{algorithm}[hbt!]
\caption{Pruning algorithm for dominating set}\label{alg:prune}
\DontPrintSemicolon
\KwData{$\mathcal{G}=(\mathcal{V, E})$ and a dominating set of $\mathcal{G}$, $S=\left\{v_{k_1}, v_{k_2}, \dots, v_{k_{|S|}} \right\} \subseteq \mathcal{V}$}
\KwResult{The pruned dominating set, $S'$}
\Begin{
    $S' \gets S$\;
    \For{$i \gets |S|$ \KwTo $1$}{
        \If{$S'\setminus\{v_{k_i}\}$ dominates $\mathcal{G}$}{
            $S'\gets S'\setminus\{v_{k_i}\}$
        }
    }
    \KwRet{$S'$}\; 
}
\end{algorithm}

As mentioned in Section \ref{sec:ig}, we also test a variant of the IG algorithm that uses the heuristics learned by the GCN model for its $InitialSolution$ and $Reconstruction$ phases. Specifically, we use the heuristic given by the first probability map $h^1_f(v)$ for the $InitialSolution$ procedure, and we then cycle through the $m$ different heuristics for the $Reconstruction$ phase of each iteration. That is, we use $h^1_f(v)$ on the first iteration, $h^2_f(v)$ on the second iteration, and so on. Once $h^m_f(v)$ is reached, we cycle back to $h^1_f(v)$ on the next iteration and repeat. In what follows, we will refer to this setup simply as the ``IG-GCN'' algorithm and the traditional IG algorithm (with the classical greedy heuristic $h_g(v)$ as described in Section \ref{sec:ig}) as the ``IG'' algorithm. In the subsequent section, we delve into the specifics of our experimental setup and conduct a comprehensive comparison of numerical results.

%% file: sections/results.tex
\section{Experimental Evaluation}\label{sec:results}

In this section, we outline the details of our experimental setup and present the numerical results of our evaluation. For our experiments, we train the GCN architecture as described in Section \ref{sec:gcn} with the following network parameters. We use $L=20$ graph convolutional layers and set the number of feature channels in each layer as $C^l = 32$ for all $l = 1, 2, \dots, L$. Therefore, the model outputs $m=C^L=32$ output probability maps for any given input graph. We train the GCN over $250$ training epochs, with a learning rate of $0.001$. Furthermore, we use $\beta = 0.2$ and $\Delta = 200$ for the IG and IG-GCN procedures, as these values were experimentally tuned in \cite{iterative}.

\begin{figure}[!b]
     \centering
     \begin{subfigure}[b]{0.49\columnwidth}
         \centering
         \includegraphics[width=\columnwidth]{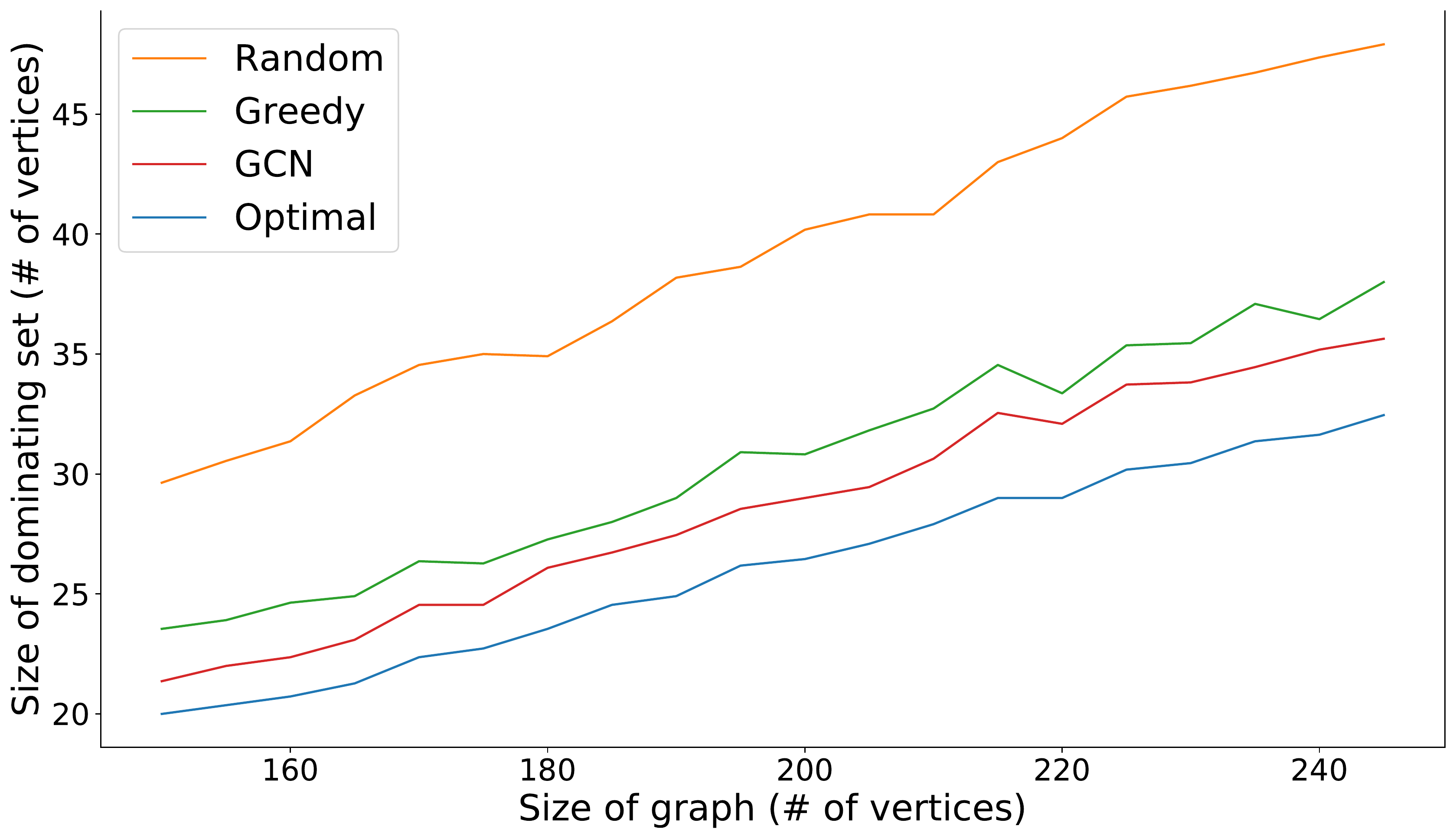}
         \caption{Comparison of the GCN, Greedy, and Random algorithms, representing the three different heuristics tested with Algorithm \ref{alg:heuristic}.}
         \label{fig:test-a}
     \end{subfigure}
     \hfill
     \begin{subfigure}[b]{0.49\columnwidth}
         \centering
         \includegraphics[width=\columnwidth]{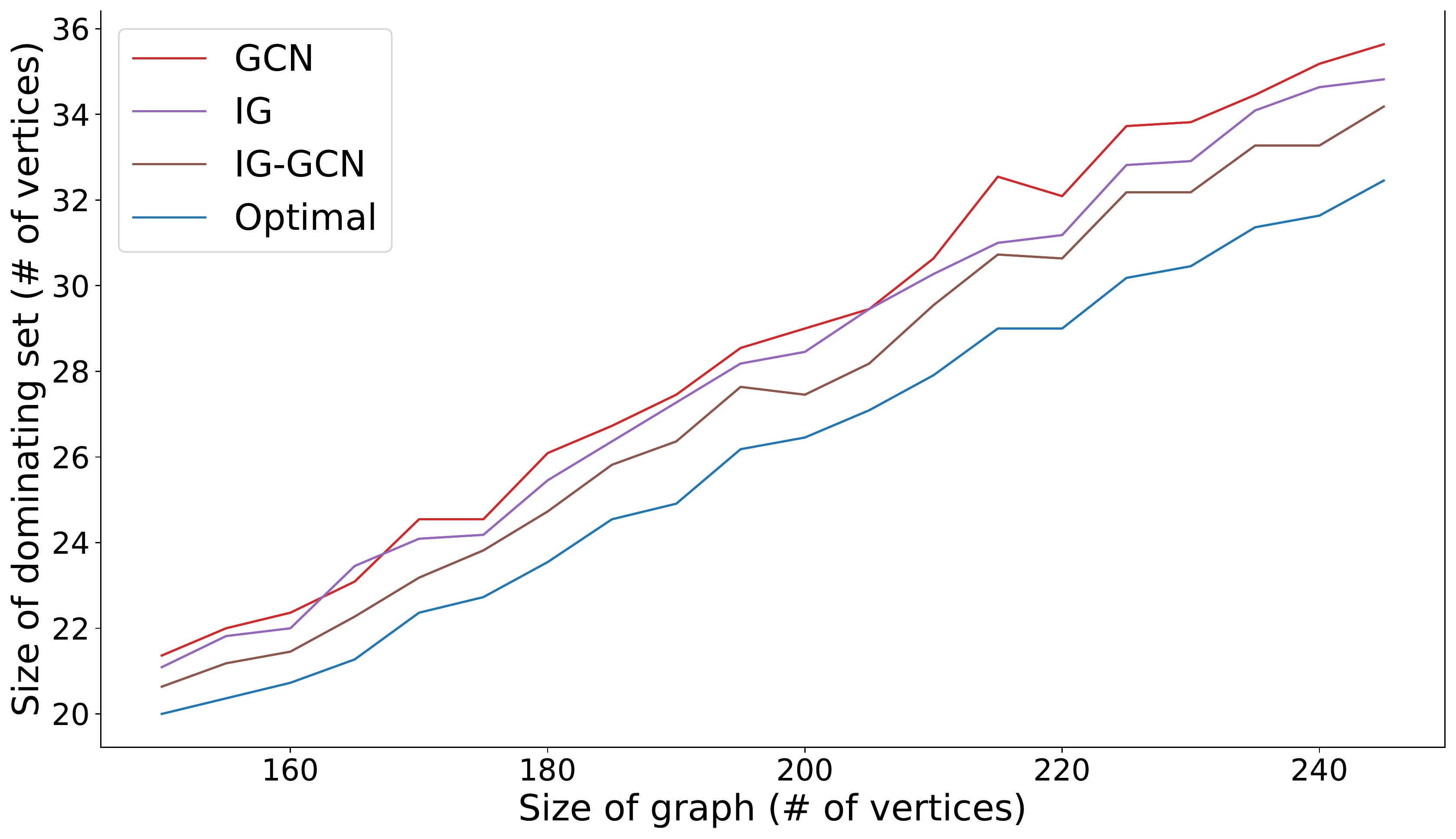}
         \caption{Comparison of the traditional IG algorithm with the IG-GCN algorithm. We reproduce the curve for the GCN algorithm here for comparison.}
         \label{fig:test-b}
     \end{subfigure}
     
    \caption{Comparison of all procedures on hold-out graphs from dataset of random binomial graphs generated using the Erd\H{o}s–R\'{e}nyi model.}
    \label{fig:test}
    \vspace{.2in}
\end{figure}

As mentioned in Section \ref{sec:gcn}, we use $1122$ graphs $(83.2\%)$ from our generated dataset of synthetic graphs to train the GCN network, leaving the remaining $227$ graphs $(16.8\%)$ for testing the performance once the training is complete. The results, in terms of the size of the dominating sets returned by each algorithm on these test graphs, are displayed in Figure \ref{fig:test}. 

Our results clearly demonstrate that the GCN algorithm effectively constructs dominating sets that consistently outperform the Greedy algorithm, resulting in smaller set sizes. Figure \ref{fig:test-b} provides a comprehensive comparison between the standard IG algorithm and the IG-GCN algorithm, reaffirming the significant performance improvement achieved by replacing the classical greedy heuristic with the GCN-based heuristic. Moreover, we present the sizes of the optimal solutions for reference, while plotting the sizes of the dominating sets returned by the GCN algorithm in both figures to facilitate a thorough comparison.

\begin{figure}[tb]
     \centering
     \begin{subfigure}[b]{0.49\columnwidth}
         \centering
         \includegraphics[width=\columnwidth]{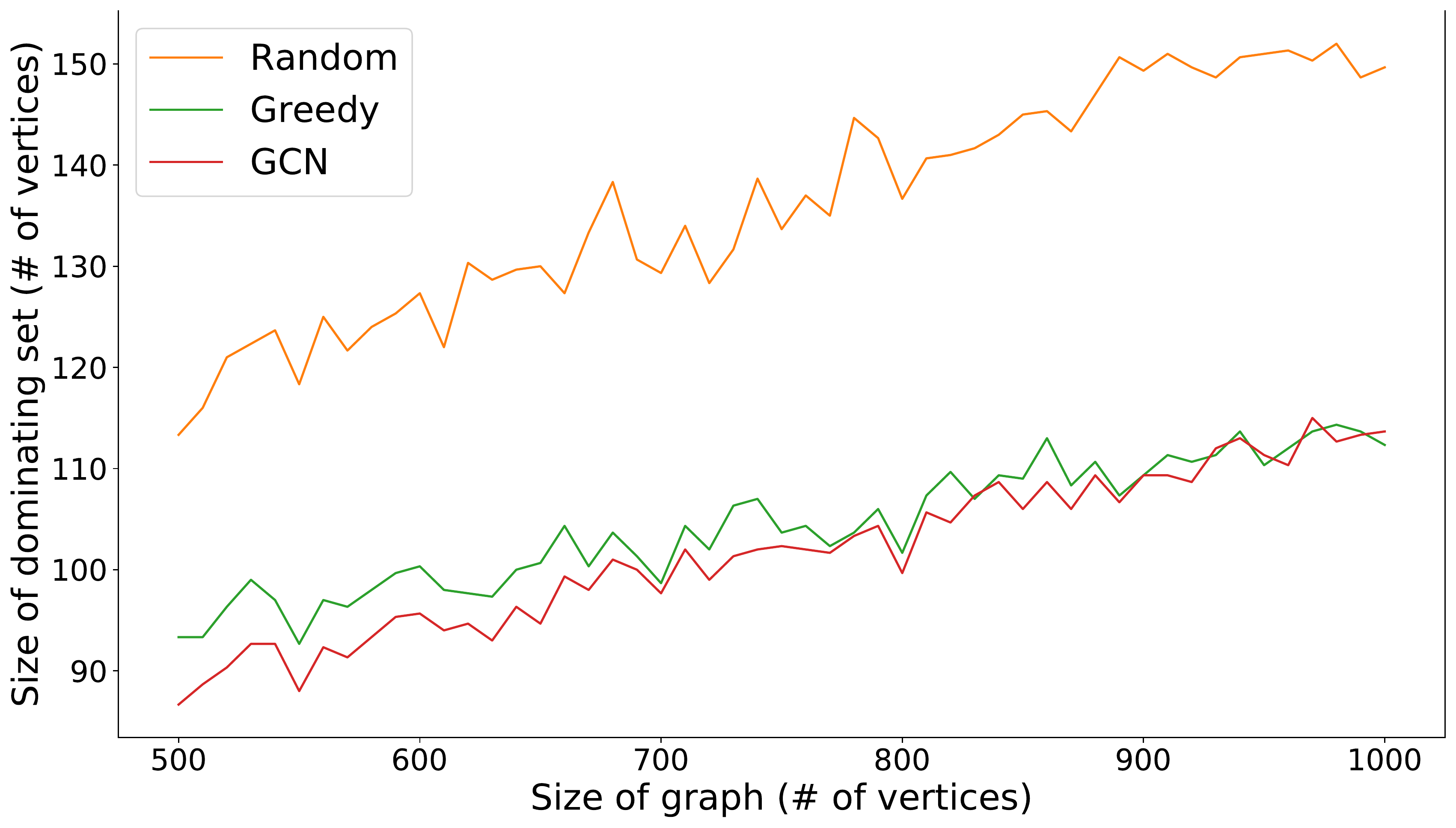}
         \caption{Comparison of the GCN, Greedy, and Random algorithms, representing the three different heuristics tested with Algorithm \ref{alg:heuristic}.}
         \label{fig:extend-a}
     \end{subfigure}
     \hfill
     \begin{subfigure}[b]{0.49\columnwidth}
         \centering
         \includegraphics[width=\columnwidth]{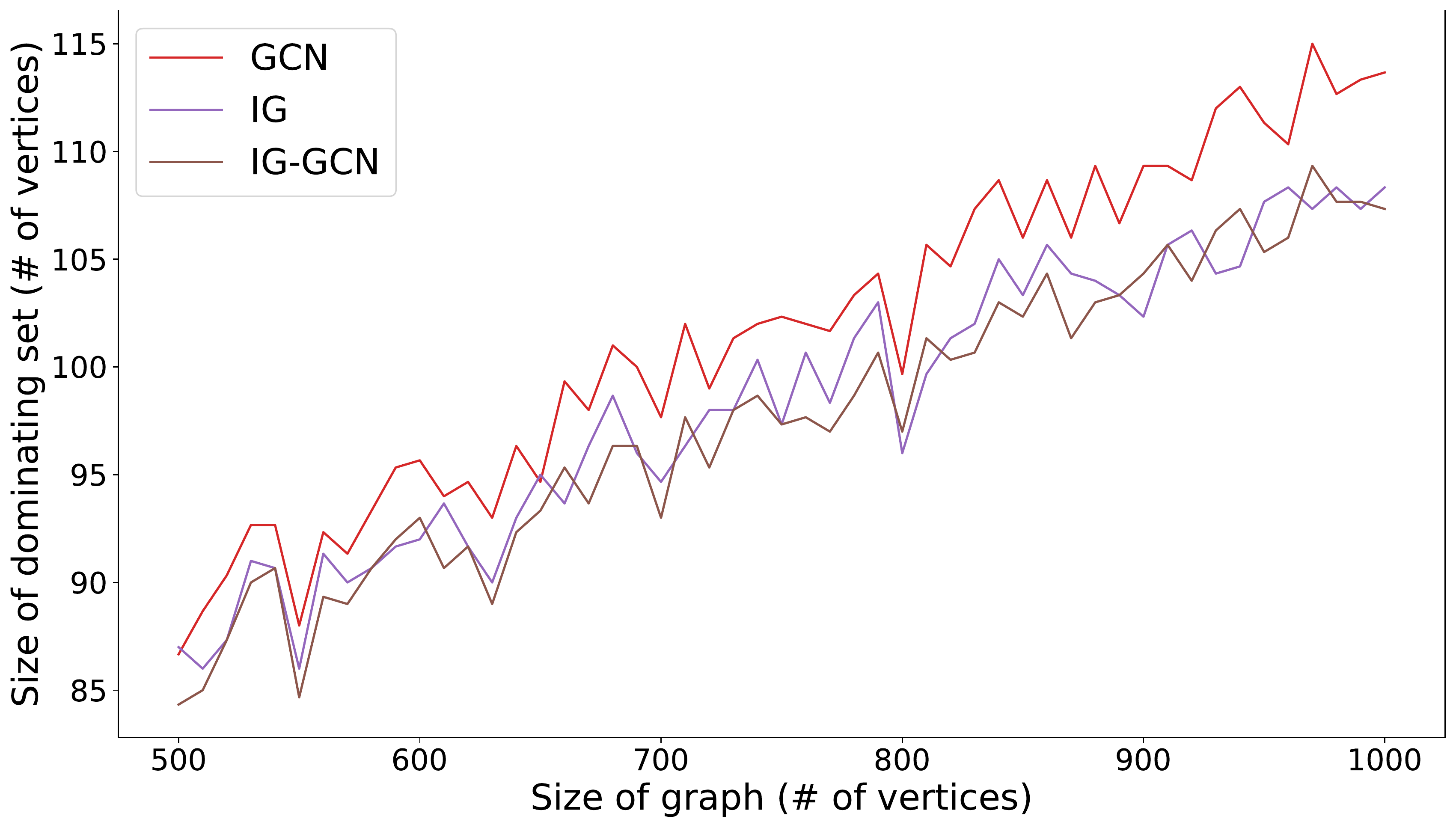}
         \caption{Comparison of the traditional IG algorithm with the IG-GCN algorithm. We reproduce the curve for the GCN algorithm here for comparison.}
         \label{fig:extend-b}
     \end{subfigure}
     
    \caption{Comparison of all procedures on higher order random binomial graphs generated using the Erd\H{o}s–R\'{e}nyi model.}
    \label{fig:extend}
\end{figure}

We repeat this experimental procedure on larger synthetically-generated graphs of $500$ to $1000$ vertices. Once again, we randomly generate these graphs using the Erd\H{o}s–R\'{e}nyi model as before. Note that on these larger instances, it is computationally intractable to compute the optimal MDS, and hence the optimal solution sizes for these graphs are unknown. However, we present the results of using the GCN and other baseline heuristics in Figure \ref{fig:extend}. As shown in Figure \ref{fig:extend-a}, even for graphs larger than the instances on which the GCN-based architecture is trained, the GCN algorithm is able to outperform the Greedy algorithm in general. In Figure \ref{fig:extend-b}, we observe that the IG-GCN algorithm is generally able to outperform the standard IG algorithm, but this performance margin is smaller than the corresponding margin on smaller order graphs. 


\begin{figure}[tb]
     \centering
     \begin{subfigure}[b]{0.49\columnwidth}
         \centering
         \includegraphics[width=\columnwidth]{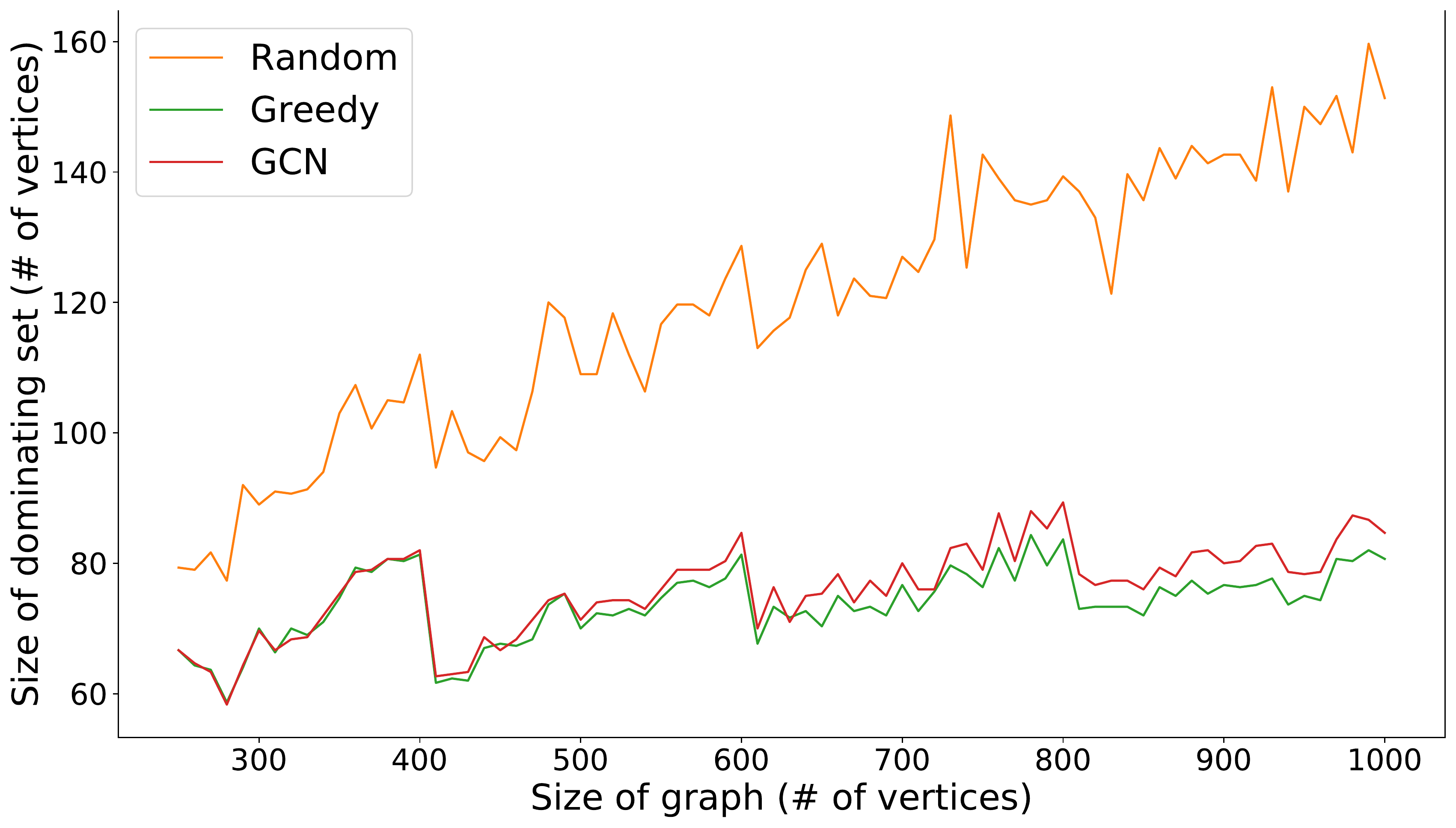}
         \caption{Comparison of the GCN, Greedy, and Random algorithms, representing the three different heuristics tested with Algorithm \ref{alg:heuristic}.}
         \label{fig:scalefree-a}
     \end{subfigure}
     \hfill
     \begin{subfigure}[b]{0.49\columnwidth}
         \centering
         \includegraphics[width=\columnwidth]{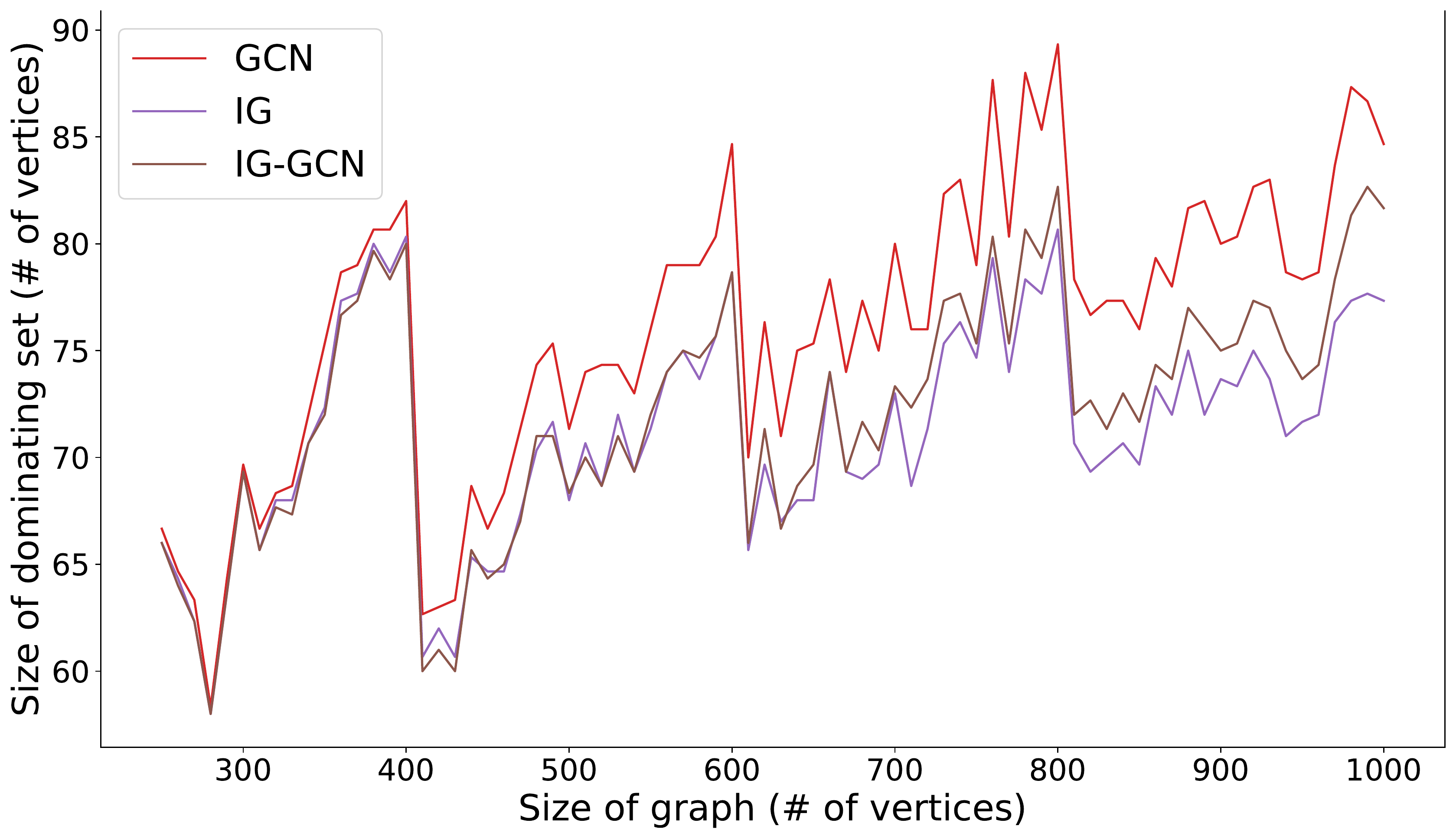}
         \caption{Comparison of the traditional IG algorithm with the IG-GCN algorithm. We reproduce the curve for the GCN algorithm here for comparison.}
         \label{fig:scalefree-b}
     \end{subfigure}
     
    \caption{Comparison of all procedures on random scale-free networks generated with the Barab\'{a}si–Albert model.}
    \label{fig:scalefree}
\end{figure}

We also test the performance of these algorithms with graphs generated using the Barab\'{a}si-Albert model in order to determine the extent to which the model of randomness used affects the performance of the GCN-based algorithms. While the Erd\H{o}s–R\'{e}nyi model is used to generate random binomial graphs, the Barab\'{a}si–Albert model uses a preferential attachment mechanism to generate random scale-free networks. The results on these graphs are given in Figure \ref{fig:scalefree}. We find that in this case, the Greedy and IG algorithms are able to outperform the GCN and IG-GCN algorithms. This is to be expected since the Greedy and IG algorithms select vertices based on the number of non-dominated neighbors of a given vertex, making them more likely to perform well in the setting of scale-free networks.

Finally, we conduct a series of experiments on a multitude of real-world graph datasets to assess the practicality of the GCN and IG-GCN methods on graphs that are derived from real-world phenomena. We provide a summary of information on these datasets in Table \ref{tab:rw-datasets}, and the results are presented in Table \ref{tab:results}. We aim to include a diverse range of datasets with varying graph sizes and settings, including biological networks, social networks, and computer vision. Our findings consistently demonstrate that the IG-GCN method achieves state-of-the-art performance, surpassing all other existing MDS algorithms across all datasets.

\begin{table*}[!b]
    \centering
    \begin{tabular}{p{0.25\textwidth} p{0.2\textwidth} p{0.2\textwidth} p{0.2\textwidth}}
        \toprule
        Dataset & Number of Graphs & Mean Number\newline of Vertices & Mean MDS Size \\
        \toprule
        
        BZR \cite{DHFR_BZR} & 405 & 35.751 & 13.114 \\ \midrule 
        dblp\_ct1 \cite{facebook} & 755 & 52.87 & 8.325 \\ \midrule 
        DD \cite{DD1, DD2} & 1176 & 276.382 & 50.392 \\ \midrule 
        DHFR \cite{DHFR_BZR} & 756 & 42.427 & 13.897 \\ \midrule
        facebook\_ct2 \cite{facebook} & 995 & 95.723 & 19.982 \\ \midrule
        FIRSTMM\_DB \cite{FIRSTMM, MSRC} & 41 & 1377.268 & 279.829 \\ \midrule 
        github\_stargazers \cite{github}  & 12725 & 113.795 & 17.402 \\ \midrule 
        MSRC\_21 \cite{MSRC} & 563 & 77.52 & 13.732 \\ \midrule 
        NCI1 \cite{NCI1, DD2} & 4110 & 29.761 & 9.776 \\ \midrule 
        OHSU \cite{OHSU} & 79 & 82.013 & 20.494 \\ \midrule
        REDDIT-MULTI-5K \cite{reddit} & 4999 & 508.507 & 97.941 \\ \midrule
        SYNTHETICnew \cite{synthetic} & 300 & 100.00 & 23.977 \\ \bottomrule
    \end{tabular}
    \caption{Real-world datasets used in experimental evaluation.}
    \label{tab:rw-datasets}
\end{table*}

\begin{table*}[!b]
    \centering
    \begin{tabular}{p{0.2\textwidth} p{0.13\textwidth} p{0.13\textwidth} p{0.13\textwidth} p{0.13\textwidth} p{0.13\textwidth}}
        \toprule
        Dataset & Random & Greedy & GCN & IG & IG-GCN \\
        \toprule

        BZR & 15.06 (15.31) & 13.13 (0.16) & \textbf{13.11 (0.0)} & \textbf{13.11 (0.0)} & \textbf{13.11 (0.0)} \\ \midrule
        dblp\_ct1 & 12.14 (47.39) & 8.33 (0.05) & \textbf{8.32 (0.0)} & \textbf{8.32 (0.0)} & \textbf{8.32 (0.0)} \\ \midrule
        DD & 65.89 (30.52) & 57.70 (14.12) & 53.56 (5.77) & 53.75 (6.36) & \textbf{52.48 (3.45)} \\ \midrule
        DHFR & 16.96 (22.43) & 13.91 (0.11) & \textbf{13.90 (0.0)} & \textbf{13.90 (0.0)} & \textbf{13.90 (0.0)} \\ \midrule
        facebook\_ct2 & 28.36 (44.57) & 19.99 (0.04) & \textbf{19.98 (0.0)} & \textbf{19.98 (0.0)} & \textbf{19.98 (0.0)} \\ \midrule
        FIRSTMM\_DB & 363.95 (29.56) & 334.32 (20.17) & 311.00 (11.00) & 304.00 (8.54) & \textbf{302.1~(7.47)} \\ \midrule
        github\_stargazers & 21.77 (30.32) & 17.43 (0.16) & 17.49 (0.13) & 17.40 (0.01) & \textbf{17.40 (0.0)} \\ \midrule
        MSRC\_21 & 18.76 (36.96) & 15.21 (10.87) & 14.19 (3.37) & 14.25 (3.84) & \textbf{13.80 (0.52)} \\ \midrule
        NCI1 & 11.64 (19.56) & 10.21 (4.32) & 9.80 (0.25) & 9.81 (0.36) & \textbf{9.78 (0.03)} \\ \midrule
        OHSU & 24.76 (19.66) & 21.34 (3.40) & 20.67 (0.68) & 20.65 (0.51) & \textbf{20.51 (0.05)} \\ \midrule        
        REDDIT-MULTI-5K & 120.55 (21.75) & 98.13 (0.20) & 98.09 (0.08) & \textbf{97.94 (0.0)} & \textbf{97.94 (0.0)} \\ \midrule
        SYNTHETICnew & 31.61 (31.88) & 27.02 (12.73) & 24.90 (3.87) & 24.86 (3.71) & \textbf{24.35 (1.57)} \\ \bottomrule
    \end{tabular}
    \caption{Comparison of performance of testing procedures on real-world datasets. We report the mean dominating set size given by each testing procedure per dataset, with the mean deviation from the optimal MDS size given as a percent in parenthesis (we omit the parenthesis sign for brevity). The best performing testing procedure(s) per dataset is given in bold.}
    \label{tab:results}
\end{table*}

%% file: sections/conclusion.tex
\section{Conclusion}\label{sec:conclusion}
In this paper, we have presented a novel approach to address the NP-hard problem of computing minimum dominating sets. Leveraging the capabilities of graph convolutional networks (GCNs), we have introduced a data-driven methodology that surpasses conventional greedy or random heuristics. Our experimental results demonstrate that our GCN approach exhibits remarkable performance, yielding near-optimal solutions for both synthetic and real-world datasets. Notably, our model showcases exceptional generalization capabilities, extending its effectiveness to graphs of higher order compared to its training set. Furthermore, our research shows that the GCN model can effectively apply its learned knowledge to real-world graphs, despite being trained exclusively on synthetically-generated random graphs. This underscores the robustness and adaptability of our proposed methodology. Additionally, by integrating the GCN-based heuristics into the iterative greedy (IG) framework, we have achieved state-of-the-art performance in the computation of dominating sets. This breakthrough not only highlights the effectiveness of our approach but also paves the way for advancements in solving complex combinatorial optimization problems.